\documentclass[journal]{IEEEtran}
\usepackage{amsmath,amsfonts}
\usepackage{algorithmic}
\usepackage{algorithm}
\usepackage{array}
\usepackage[caption=false,font=normalsize,labelfont=sf,textfont=sf]{subfig}
\usepackage{textcomp}
\usepackage{stfloats}
\usepackage{url}
\usepackage{verbatim}
\usepackage{graphicx}
\usepackage{cite}

\usepackage{booktabs}
\usepackage {multirow}
\usepackage{xcolor}
\begin{document}

\title{Advancing Real-World Parking Slot Detection with Large-Scale Dataset and Semi-Supervised Baseline}

\author{Zhihao Zhang, Chunyu Lin, \IEEEmembership{Member,~IEEE}, Lang Nie, Jiyuan Wang, Yao Zhao, \IEEEmembership{Fellow, IEEE}


\thanks{Zhihao Zhang, Chunyu Lin, Lang Nie, Jiyuan Wang and Yao Zhao are with the Institute of Information
Science, Beijing Jiaotong University, Beijing 100044, China, and also with
Visual Intelligence +X International Cooperation Joint Laboratory of MOE,
Beijing 100044, China (zhihao.zhang@bjtu.edu.cn; cylin@bjtu.edu.cn; nielang@bjtu.edu.cn; wangjiyuan@bjtu.edu.cn; yzhao@bjtu.edu.cn).}
\thanks{This work was supported by the National Natural Science Foundation of China (NSFC) under Grant (U2441242,62172032) (Corresponding author: Chunyu Lin)}
}

\markboth{Journal of \LaTeX\ Class Files,~Vol.~14, No.~8, August~2021}%
{Shell \MakeLowercase{\textit{et al.}}: A Sample Article Using IEEEtran.cls for IEEE Journals}


\maketitle

\begin{abstract}
As automatic parking systems evolve, the accurate detection of parking slots has become increasingly critical. This study focuses on parking slot detection using surround-view cameras, which offer a comprehensive bird’s-eye view of the parking environment.
However, the current datasets are limited in scale, and the scenes they contain are seldom disrupted by real-world noise (\textit{e.g.}, light, occlusion, etc.). Moreover, manual data annotation is prone to errors and omissions due to the complexity of real-world conditions, significantly increasing the cost of annotating large-scale datasets. To address these issues, we first construct a large-scale parking slot detection dataset (named CRPS-D), which includes various lighting distributions, diverse weather conditions, and challenging parking slot variants. Compared with existing datasets, the proposed dataset boasts the largest data scale and consists of a higher density of parking slots, particularly featuring more slanted parking slots. 
Additionally, we develop a semi-supervised baseline for parking slot detection, termed SS-PSD, to further improve performance by exploiting unlabeled data. 
To our knowledge, this is the first semi-supervised approach in parking slot detection, which is built on the teacher-student model with confidence-guided mask consistency and adaptive feature perturbation.
Experimental results demonstrate the superiority of SS-PSD over the existing state-of-the-art (SoTA) solutions on both the proposed dataset and the existing dataset. Particularly, the more unlabeled data there is, the more significant the gains brought by our semi-supervised scheme. The relevant source codes and the dataset have been made publicly available at https://github.com/zzh362/CRPS-D.
\end{abstract}

\begin{IEEEkeywords}
Parking slot detection, large-scale dataset, semi-supervised baseline.
\end{IEEEkeywords}

\section{Introduction}
\begin{figure}[t]
\centering
\includegraphics[width=0.95\linewidth]{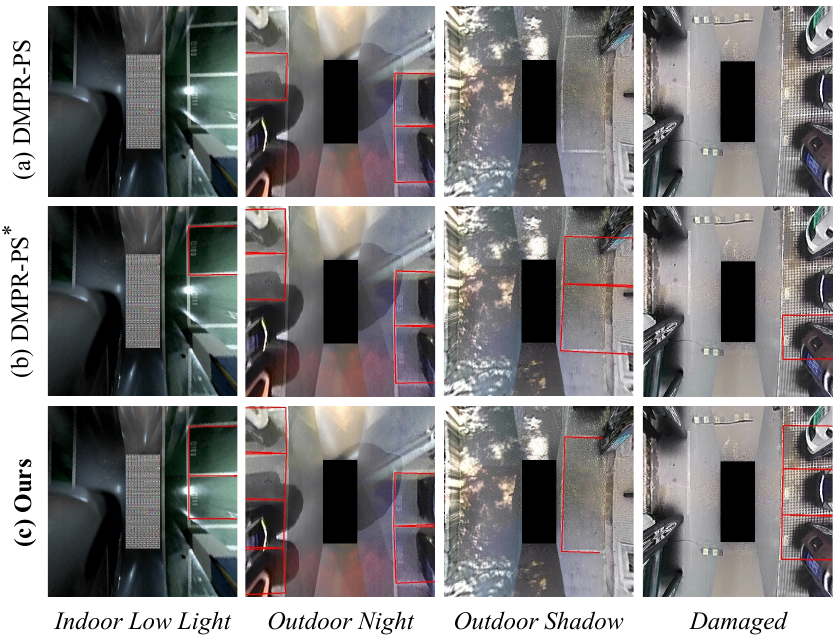}
\caption{\textbf{Parking slot detection on real-world challenging scenes.} (a) The results of DMPR-PS~\cite{huang2019dmpr} with the pre-trained model. (b) The results of DMPR-PS trained on the proposed CRPS-D dataset with 1/12 labeled data. (c) The results of our semi-supervised baseline trained on the CRPS-D dataset with 1/12 labeled data and 11/12 unlabeled data.}
\label{fig1}
\end{figure}

\IEEEPARstart{A}{utomatic} parking technology plays an indispensable role in the field of autonomous driving.
Accurate parking slot detection is a critical first step in autonomous parking systems, directly influencing subsequent vehicle planning. In this context, surround-view camera systems have emerged as the preferred solution, providing a comprehensive bird’s-eye view of the vehicle’s surroundings and enabling more effective detection of parking slots.
Traditional parking slot detection methods~\cite{jung2006parking, jung2009uniform, jung2006structure, suhr2012fully, lee2018probabilistic, suhr2013sensor} heavily relied on hand-crafted features like corners and lines, which are far from providing satisfactory accuracy, especially in scenes where geometric features are insufficient.
With the development of various deep learning-based object detection techniques, learning-based parking slot detection algorithms have now superseded conventional methods. 
The introduction of the first learning-based algorithm, DeepPS~\cite{zhang2018vision}, along with the benchmark dataset ps2.0, has catalyzed progress in this field. Researchers have since focused on improving detection accuracy through various approaches, such as marking point regression (DMPR-PS\cite{huang2019dmpr}), Graph Neural Networks (GNNs\cite{min2021attentional}), and transformer-based~\cite{bui2023transformer} methods, etc. 
These algorithms have recently achieved SoTA performance on the benchmark dataset ps2.0, with an accuracy rate of over 99\%. However, in reality, parking scenes often present greater complexity and difficulty. When these solutions are applied to real-world automatic parking scenarios, as shown in Fig. \ref{fig1}(a), they frequently fail to meet performance expectations.
Consequently, it is urgent to build a larger-scale, challenging, and realistic dataset to advance real-world parking slot detection and improve the assessment of algorithm viability.

To fulfill this requirement, we gathered and meticulously annotated a large-scale real-world dataset called CRPS-D for parking slot detection. 
It comprises a variety of parking slots, such as above-ground parking lots, underground parking structures, street parking areas, and so on, aiming to cover a wide range of real-world parking scenes encountered in urban environments.
The data was collected across different times and under various weather conditions, including daytime, nighttime, sunny, cloudy, and rainy. 
To our knowledge, the proposed dataset surpasses all publicly available datasets (\textit{e.g.}, ps2.0~\cite{zhang2018vision} and SNU~\cite{do2020context}) in scale by a significant margin. 
Moreover, compared with these datasets, CRPS-D exhibits a higher concentration of parking slots, a larger quantity of slanted parking slot occurrences, and more intricate road scenes. These factors collectively increase the complexity of detecting parking slots, providing a comprehensive and challenging benchmark that can drive advancements in real-world parking slot detection.

On the other hand, we observed that existing parking slot detection solutions heavily depend on fully supervised learning schemes, which require extensive manual annotation. 
Although semi-supervised learning has been widely used in tasks such as object detection and keypoint estimation, it has not been extensively explored in the context of parking slot detection. Directly applying these methods to parking slot detection, however, presents several unique challenges. Unlike object detection, which relies on bounding boxes for coarse localization, parking slots are defined by a set of keypoints. These keypoints are subject to strict geometric constraints, making pseudo-label generation and consistency enforcement particularly challenging.
To address this issue, we propose the first semi-supervised model designed to alleviate the need for expensive labels. This model is built upon the teacher-student framework~\cite{tarvainen2017mean} and incorporates customized confidence-guided mask (CGM) consistency and adaptive feature perturbation (Adaptive-VAT).
Concretely, the teacher model is an exponential moving average (EMA) of the student model and helps guide the student model through consistency regularization. However, constraining potentially incorrect predictions to be consistent can have negative impacts. To mitigate this, we propose using a trainable confidence map to assign different weights to various areas and mask low-confidence regions. This approach formulates a confidence-guided mask consistency constraint for unlabeled data.
Additionally, considering that the goal of the teacher-student model is to encourage the student model to produce stable outputs under different perturbations (\textit{i.e.},  T-VAT~\cite{liu2022perturbed}), we design Adaptive-VAT, an adaptively selective feature perturbation mechanism, to generate stronger but reasonable adversarial noise. 
Our Adaptive-VAT selectively adopts either the teacher or student model based on their resistance robustness to perturbations, generating challenging noise to promote the effective training of the student model.

To sum up, our contributions center around the following:
\begin{itemize}
\item We build \textbf{CRPS-D}, the largest benchmark dataset for real-world parking slot detection. Compared with existing datasets, CRPS-D potentially advances the field with its larger data size, denser instances, and more challenging scenes.

\item To reduce reliance on expensive annotations, we propose the first semi-supervised baseline (termed \textbf{SS-PSD}) for parking slot detection, which is built on the teacher-student model with confidence-guided mask consistency and adaptive feature perturbation.

\item The proposed solution achieves SOTA performance on both publicly available datasets and our CRPS-D benchmark dataset, particularly with larger amounts of unlabeled data.
\end{itemize}

\begin{figure*}[!t]
\centering
\includegraphics[width=0.95\textwidth]{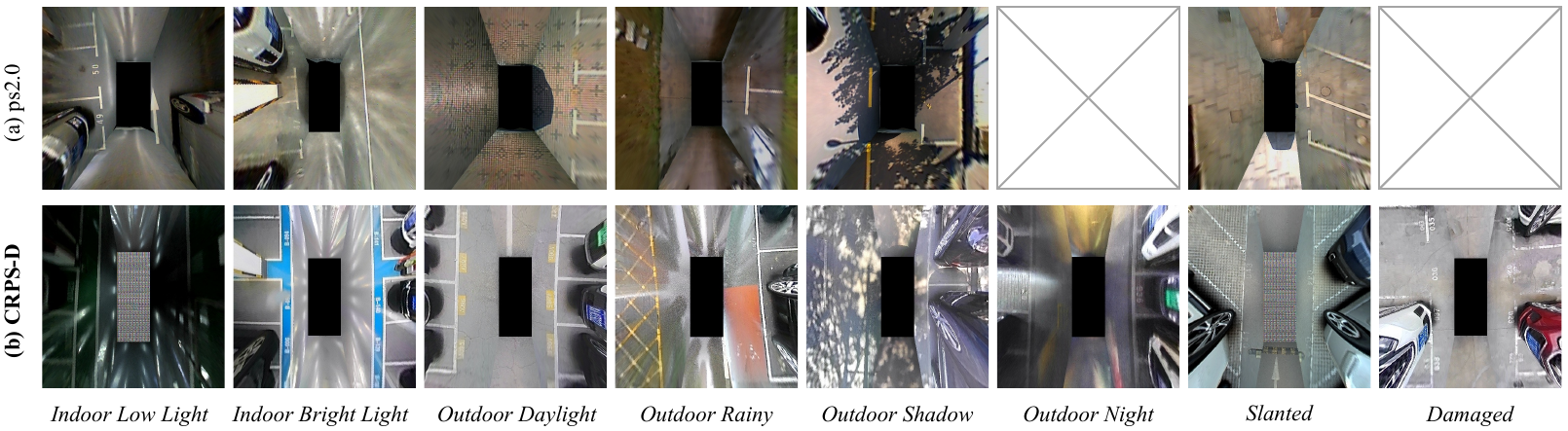}
\caption{\textbf{Example images.} The scenes are from (a) the publicly available dataset ps2.0, and (b) the proposed CRPS-D dataset.}
\label{fig2}
\end{figure*}

\section{Related Work}
In this section, we provide a brief overview of related benchmark datasets and deep learning methods for parking slot detection, as well as semi-supervised learning (SSL) methods.

\subsection{Benchmark Datasets.}
Two widely used benchmark datasets are employed for training and evaluating parking slot detection tasks, the ps2.0~\cite{zhang2018vision} dataset and the SNU~\cite{do2020context} dataset. 
The ps2.0 dataset is the first benchmark dataset for parking slot detection research. It includes 9,827 images for training and an additional 2,338 images for testing. The dataset is compiled by integrating images captured from four fish-eye cameras, providing a comprehensive bird’s-eye view perspective. The SNU dataset comprises 18,299 images for training and an additional 4,518 images for testing. The images were collected using fish-eye cameras mounted on the sides of vehicles without post-processing to stitch them together. 
Researchers predominantly rely on these datasets to evaluate the effectiveness of their detection algorithms.

\subsection{Deep Learning Methods.}
Methods for parking slot detection based on deep learning can mainly be subdivided into four categories~\cite{wong2023review}: object detection~\cite{zhang2018vision, zinelli2019deep, do2020context, li2020vacant, li2020parking, suhr2021end, huang2022visual, zheng2022center, wang2022detps, bui2023cnn, lee2023keps, bui2023one}, image segmentation~\cite{wu2018vh, jiang2019dfnet, jang2019semantic, jian2020vision, yu2020spfcn, zhou2022passim, lai2022semantic}, regression~\cite{huang2019dmpr}, GNNs~\cite{min2021attentional}, and transformer-based method~\cite{bui2023transformer}. 
DeepPS~\cite{zhang2018vision} is the first DCNN-based method for detecting parking slots, using YOLOv2~\cite{redmon2017yolo9000} to identify marking points and a custom network to determine slot orientation, with final slot locations inferred from these detections.
Following this, methods such as VPS-Net~\cite{li2020vacant} have also been developed for parking slot detection. 
Applying segmentation techniques, Jian and Lin~\cite{jian2020vision} cascaded two U-Nets: the first segments slot lines, while the second segments corner points. Post-processing then filters out invalid points and identifies valid parking slot coordinates.
In regression-based methods, the focus is on predicting the coordinates and orientation of marking points. For example, DMPR-PS~\cite{huang2019dmpr} is a custom CNN model that outputs a $16 \times 16 \times 6$ matrix representing directional marking points. Subsequently, template matching is employed to classify whether these points form valid entrance lines.
Min et al.~\cite{min2021attentional} were the first to apply GNNs to the parking slot detection task, leveraging their ability to model global relationships. 
Recently, Quang et al.~\cite{bui2023transformer} introduced a transformer-based method that uses fixed anchor points instead of object queries in DETR~\cite{carion2020end}, which reduces training time by focusing on predefined areas of the feature map.
In addition, several novel approaches have recently emerged, including a method relying on a single front-view camera~\cite{lungoci2024embedded}, techniques using fisheye images~\cite{fu2024parkscape, musabini2024enhanced}, and a zero-shot framework leveraging large visual models~\cite{zhai2024sam}.

Beyond vision-based methods, V2X technologies have been explored for improved localization in challenging environments. For instance, CV2X-LOCA~\cite{huang2024toward} leverages RSU-based cooperation to achieve lane-level accuracy in GPS-denied areas, which could offer valuable support for parking slot detection in settings such as tunnels or underground garages.

\subsection{Semi-Supervised Learning Methods.}
Existing SSL approaches are based on three main assumptions~\cite{tarvainen2017mean}: smoothness (similar images are likely to share labels), low-density (decision boundaries avoid regions of high-density in the feature space), and manifold (instances on the same low-dimensional manifold tend to have the same label).
SSL methods can broadly be categorized into two groups: pseudo-label based~\cite{berthelot2019mixmatch, arazo2020pseudo, sohn2020fixmatch} and consistency based~\cite{polyak1992acceleration, laine2016temporal, tarvainen2017mean, liu2022perturbed}. 
Pseudo-label methods often exhibit lower accuracy than consistency-based methods, likely due to the exclusion of some unlabeled data during training, which diminishes their generalization capacity.
Consistency-based SSL methods enforce prediction consistency across perturbed unlabeled images, demonstrating improved performance on standard benchmarks. Their efficacy depends on the prediction accuracy for unlabeled images and the nature of the perturbations applied during training.
In PS-MT~\cite{liu2022perturbed}, Liu et al. proposed a feature perturbation called T-VAT, which replaces the virtual adversarial training (VAT~\cite{miyato2018virtual}), resulting in improved model performance.

\begin{table*}[!t]
\caption{Density of parking slots. The density refers to the 
`Ratio' of the number of parking slots to the number of images.}
\centering
\begin{tabular}{lccccccccc}
\toprule
 & \multicolumn{3}{c}{ps2.0\cite{zhang2018vision}} & \multicolumn{3}{c}{SNU\cite{do2020context}} & \multicolumn{3}{c}{\textbf{Ours}} \\\cmidrule{2-4}\cmidrule{5-7}\cmidrule{8-10}%
 
    & $N_{slots}$ &$N_{images}$ & Ratio & $N_{slots}$ &$N_{images}$ & Ratio & $N_{slots}$ &$N_{images}$ & Ratio \\
\midrule
    Training & 9,476 & 9,827 & 0.96 & 48,886 & 18,299 & 2.67 & 118,057 & 29,803 & \textbf{3.96}  \\
    Testing & 2,168 & 2,336 & 0.93 & 13,545 & 4,518 & 3.00 & 24,024 & 5,936 & \textbf{4.05}  \\
    Total & 11,644 & 12,165 & 0.96 & 62,431 & 22,817 & 2.74 & 142,081 & 35,739 & \textbf{3.98}  \\
\bottomrule
\end{tabular}
\label{tab2}
\end{table*} 

\begin{table*}[!t]
\centering
\caption{Proportion of slanted parking slots. The proportion means a `Percentage' of the parking slots are slanted.}
\begin{tabular}{lccccccccc}
\toprule
 & \multicolumn{3}{c}{ps2.0\cite{zhang2018vision}} & \multicolumn{3}{c}{SNU\cite{do2020context}} & \multicolumn{3}{c}{\textbf{Ours}} \\\cmidrule{2-4}\cmidrule{5-7}\cmidrule{8-10}%
 
    & $S_{slant}$ &$S_{Total}$ & Percentage & $S_{slant}$ &$S_{Total}$ & Percentage & $S_{slant}$ &$S_{Total}$ & Percentage \\
\midrule
    Training & 316 & 9,476 & 3.33\% & 3276 & 48,886 & 6.70\% & 14,126 & 118,057 & \textbf{11.97\%}  \\
    Testing & 81 & 2,168 & 3.74\% & 1,004 & 13,545 & 7.41\% & 2,573 & 24,024 & \textbf{10.71\%}  \\
    Total & 397 & 11,644 & 3.41\% & 4,280 & 62,431 & 6.86\% & 16,699 & 142,081 & \textbf{11.75\%}  \\
\bottomrule
\end{tabular}
\label{tab4}
\end{table*} 

\section{Dataset}
Recently, parking slot detection models have shown promising performance, but they fail to generalize to real-world scenarios due to the scale-limitation and over-idealization of their training datasets. In this paper, we propose CRPS-D dataset, the largest benchmark dataset for this task, to advance parking slot detection technology and improve the feasibility assessment of algorithms.

\subsection{Data Collection and Organization}
\begin{figure}[htbp]
    \centering
    \includegraphics[width=1\linewidth]{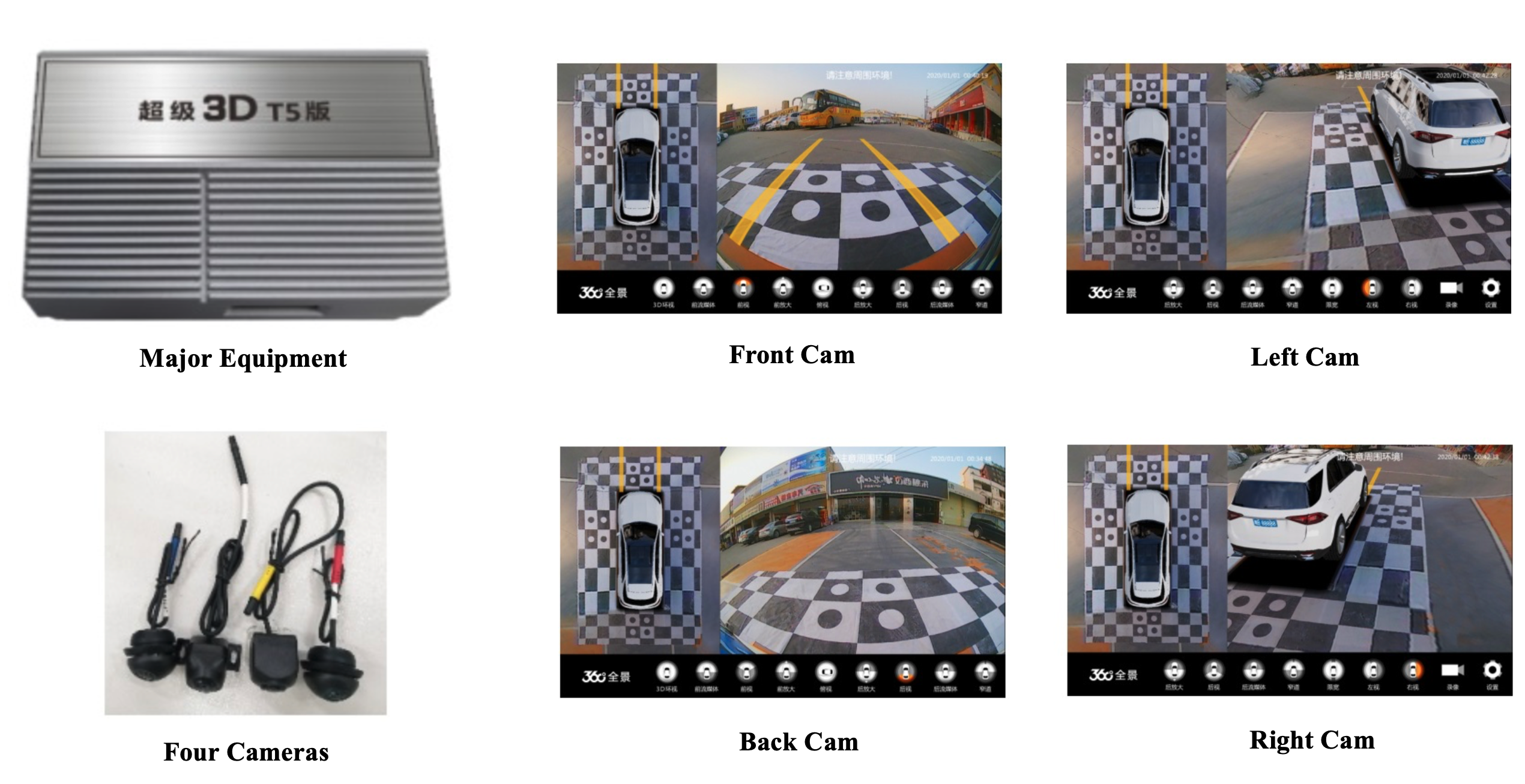}
    \caption{
    \textbf{360-degree surround view system}. The left side shows the main processing unit and four fisheye cameras, while the right side illustrates the visualization of the field of view (FOV).
    }
    \label{fig:equip}
   \end{figure}

The complete dataset consists of 35,379 images, which are divided into a training set of 29,803 images and a test set of 5,936 images. 
Each image is annotated with several marking points~\cite{huang2019dmpr} and parking slots. The marking points are characterized by their coordinates, shape, type, and angle, while the parking slots are determined by pairs of these marking points. The images are captured from a side-mirror-mounted fish-eye camera and then transformed into a bird's-eye view, with each $512 \times 512$ surround view image covering a $10m \times 10m$ area. 
Fig. \ref{fig:equip} shows the equipment we use for data collection, including the major equipment and four fisheye cameras.

For the division of the training and testing sets, we implemented a strategy that minimizes data leakage. Specifically, we organized the images based on parking lot identity, parking area, and time period, ensuring that images from the same physical location do not appear in both sets. Additionally, our dataset includes a subset of images extracted from video sequences. To avoid near-duplicate samples, we sampled frames at large intervals.

\subsection{Data Analysis}
In this part, we conduct a thorough data analysis to unravel the improvements of the proposed dataset over existing ones.

\textbf{Large-Scale Data.} 
Datasets have become increasingly crucial for current vision models (\textit{e.g.}, SAM~\cite{kirillov2023segment}), motivating us to build a larger dataset. Concretely, as shown in Table~\ref{tab1}, the total number of images and parking slots in the proposed CRPS-D dataset is 2.94 and 12.2 times that of ps2.0, and 1.57 and 2.28 times that of SNU, respectively. Here, $N$ denotes the sample number under various conditions.
Additionally, to further measure the abundance of the dataset, we also calculate the \textit{slot-image density}, expressed as 
\begin{equation}
     \textit{slot-image density} = \frac{N_{slots}}{N_{images}}.
\end{equation}
As shown in Table~\ref{tab2}, once again, the \textit{density} of our CRPS-D dataset is much higher than that of the other two datasets (3.98 \textit{v.s.} 2.74, 0.96).  This larger dataset enables the training of more extensive and complex models, better aligning with the demands of real-world applications.

\begin{table}[ht]
\centering
\caption{Numerical comparisons between the proposed dataset and publicly available datasets.}
    \begin{tabular}{llccc}
        \toprule
         Dataset  & & ps2.0\cite{zhang2018vision} & SNU\cite{do2020context} & \textbf{Ours} \\
        \midrule
             \multirow{3}{*}{$N_{images}$} & Training & 9,827 & 18,299 & 29,803  \\
              & Testing & 2,336&	4,518	&5,936   \\
              & Total & 12,165	& 22,817	& \textbf{35,739}  \\
        \midrule
            \multirow{3}{*}{$N_{slots}$ in Train Set}  & Vertical & 9,160 & 45,610 & 103,931  \\
              & Slanted & 316 &	3,276 & 14,126   \\
              & Total & 9,476 & 48,886 & \textbf{118,057}  \\
        \midrule
             \multirow{3}{*}{$N_{slots}$ in Test Set}   & Vertical & 2,087 & 12,541 & 21,451 \\
              & Slanted & 81 & 1,004 & 2,573  \\
              & Total & 2,168 & 13,545 & \textbf{24,024}   \\
        \midrule
             $N_{slots}$  & Total & 11,644 & 62,431 & \textbf{142,081} \\
        \bottomrule
    \end{tabular}
    \label{tab1}
\end{table}

\begin{figure}[ht]
\centering
\includegraphics[width=0.8\linewidth]{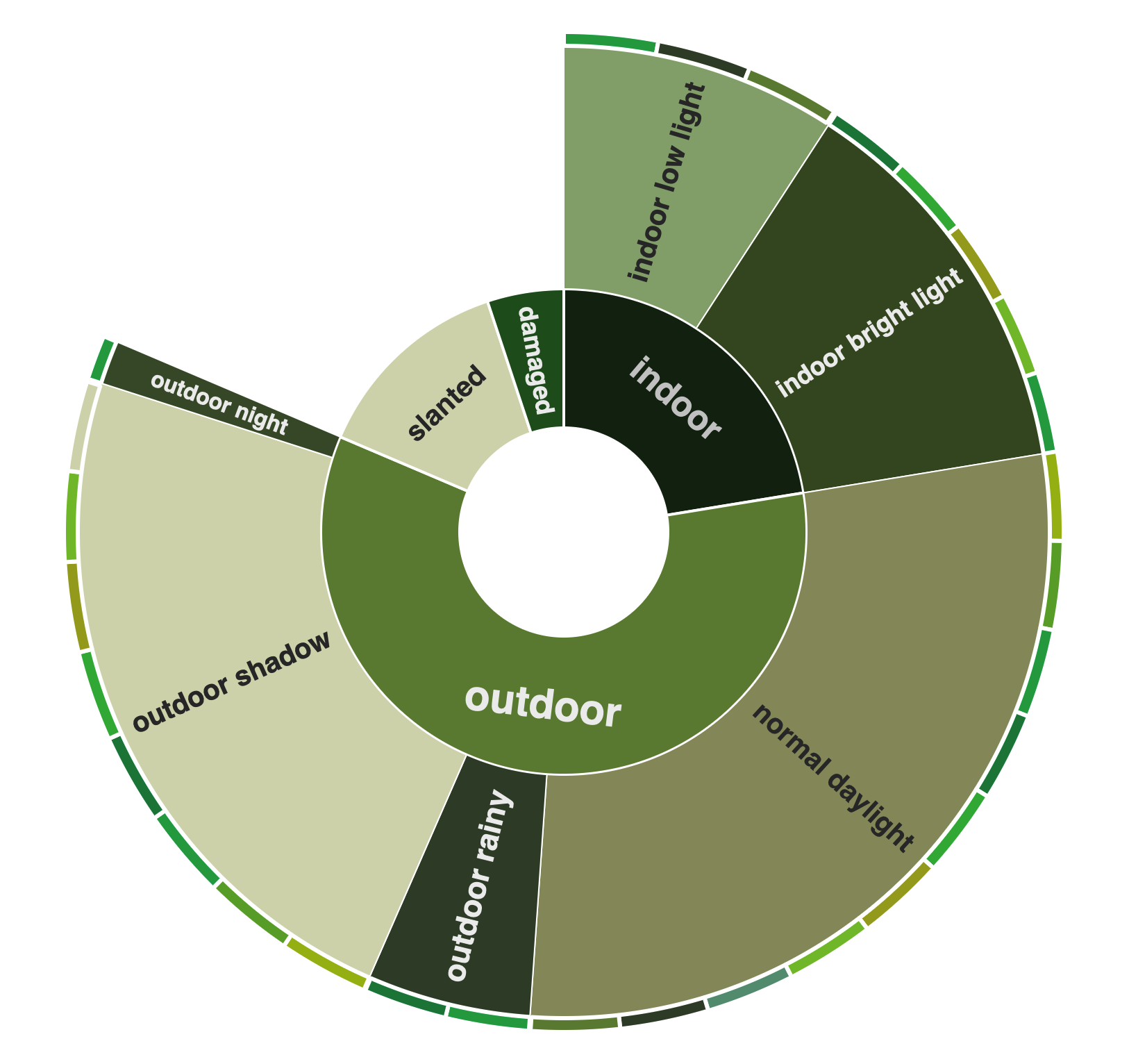}
\caption{\textbf{The scenes and corresponding hierarchical content structure} of the CRPS-D. This set is divided into four main categories: indoor, outdoor, slanted, and damaged, with indoor and outdoor scenes further subdivided.}
\label{fig_h}
\end{figure}

\textbf{Complexity of Scenes.}
To cover a wide range of parking scenes in the real world, our CRPS-D dataset contains eight subsets (\textit{i.e.}, `indoor low/bright light', `outdoor daylight/rainy/shadow/night', `slanted' and `damaged'), each corresponding to a distinct scene. As shown in Table~\ref{tab3}, the proposed dataset not only offers greater variety (\textit{i.e.}, extra `outdoor night’ and `damaged’ scenes) but also has more images in each scene compared with ps2.0. Moreover, to vividly demonstrate the strength of our dataset, we present the examples of each scene in Fig. \ref{fig2}, where it can be observed that CRPS-D is sharply more realistic than ps2.0. 
Fig. \ref{fig_h} illustrates the scenes and corresponding hierarchical content structure of CRPS-D test set. The full dataset has an almost identical scene distribution to the test set. 

\begin{table}[ht]
\centering
\setlength{\tabcolsep}{7.0mm}
\caption{Statistics for the scene data in the test set.}
\begin{tabular}{lcc}
    \toprule
    \textit{scene} & ps2.0\cite{zhang2018vision} & \textbf{Ours} \\
    \midrule
    \textit{Indoor Low Light} & \multirow{2}{*}{226} & 545  \\
    \textit{Indoor Bright Light} &  & 786  \\
    \midrule
    \textit{Outdoor Daylight} & 546 & 1,703  \\
    \textit{Outdoor Rainy} & 244 & 323  \\
    \textit{Outdoor Shadow} & 1,127 & 1,389  \\
    \textit{Outdoor Night} & / & 86  \\
    \midrule
    \textit{Slanted} & 48 & 802  \\
    \midrule
    \textit{Damaged} & / & 302  \\
    \bottomrule
\end{tabular}
\label{tab3}
\end{table} 

\textbf{Proportion of Slanted Parking Slots.}
Slanted parking slots are frequently used on narrow streets to better use road space and are commonly encountered in real-world scenes. However, slanted slots are more challenging to detect than perpendicular ones and are often neglected by other datasets. To this end, we have increased the \textit{proportion} of slanted parking slots.
As shown in Table~\ref{tab4}, the proposed CRPS-D dataset includes 11.75\% slanted parking slots, a significantly higher proportion than ps2.0 (3.41\%) and SNU (6.86\%). This provides a more realistic representation of real-world conditions and helps mitigate class imbalance issues.

\begin{figure*}[!t]
\centering
\includegraphics[width=0.95\textwidth]{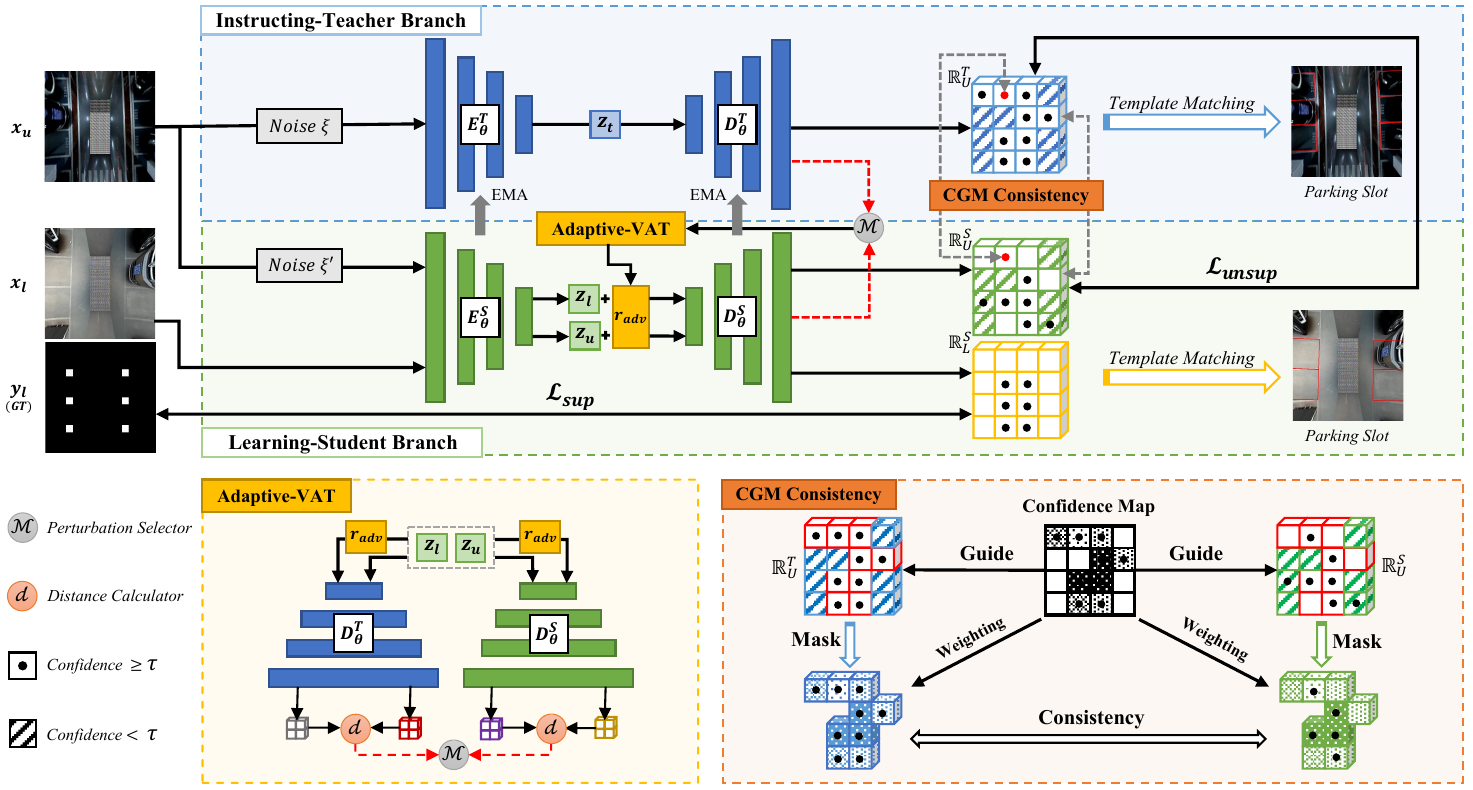}
\caption{\textbf{Overview of SS-PSD architecture.} 
The entire architecture consists of two branches: the learning-student branch that conducts the model training and the instructing-teacher branch, in which $M_{\theta}^{T}$ is updated using the EMA and the parameters of $M_{\theta}^{S}$. Positioned at the bottom left and bottom right of the figure are \textbf{Adaptive-VAT} and \textbf{CGM Consistency}, respectively.
}
\label{fig3}
\end{figure*}

\section{Methodology}
In this section, we describe the details of the SS-PSD architecture and semi-supervised learning strategy.

\subsection{Problem Formulation}
Firstly, we randomly select a small subset of datasets with corresponding labels to form a labeled dataset, denoted as $\mathcal{D}_{L}=\{(\textbf{x}_{l},\textbf{y}_{l})\}$, while the remaining images constitute the unlabeled dataset $\mathcal{D}_{U}=\{(\textbf{x}_{u})\}$. Given an image $\textbf{x}_{l} \in 512 \times 512$ and the corresponding label $\textbf{y}_{l}$, the purpose of this work is to predict a feature map $\mathbb{R} \in \mathcal{S}\times \mathcal{S} \times \mathcal{N}$, with the image being divided into $\mathcal{S}\times \mathcal{S}$ grids. Following DMPR-PS~\cite{huang2019dmpr}, the feature map $\mathbb{R}$ corresponds to the marked points in the input image, with $\mathcal{N}$ dimensions representing their respective features. Then, we apply template matching to determine whether these points can form valid entrance lines, thereby identifying the corresponding parking slots. This process can be described as follows:
\begin{equation}
    \textbf{x}_{l} \xrightarrow{M_{\theta}} \mathbb{R}^{\mathcal{S}\times \mathcal{S} \times \mathcal{N}} \xrightarrow{Template \ Matching} Parking \ Slot ,
\end{equation}
where $M_{\theta}(\cdot)$ is a model used to predict the marking point, and $\theta$ represents the model parameters. 

Distinguishing from prior works that heavily rely on expensive labeled datasets, we exploit the easily available unlabeled images $\textbf{x}_{u}$ to further improve performance.

\subsection{Network Architecture}
Fig. \ref{fig3} illustrates the architecture of the proposed semi-supervised baseline for parking slot detection (SS-PSD), which consists of an instructing-teacher branch and a learning-student branch. They have an identical network structure, denoted as $M_{\theta}=\{E_{\theta}, D_{\theta}\}$, which includes an encoder $E_{\theta}: \textbf{x}\rightarrow \textbf{z}$ and a decoder $D_{\theta}: \textbf{z}\rightarrow \mathbb{R}^{\mathcal{S} \times \mathcal{S} \times \mathcal{N}}$. 
Particularly, we define the teacher model as $M_{\theta}^{T}=\{E_{\theta}^{T}, D_{\theta}^{T}\}$ and the student model as $M_{\theta}^{S}=\{E_{\theta}^{S}, D_{\theta}^{S}\}$.

\textbf{Learning-Student Branch.}
We adopt a customized CNN model~\cite{huang2019dmpr} as our encoder-decoder architecture. In the learning-student branch, the encoder $E_{\theta}^{S}$  processes both labeled $\textbf{x}_{l}$ and unlabeled $\textbf{x}_{u}$ data, producing the latent features $\textbf{z}_{l}$ and $\textbf{z}_{u}$, which can be represented as $E_{\theta}^{S}(\textbf{x}_{l} / \textbf{x}_{u}) = \textbf{z}_{l} / \textbf{z}_{u}$. 
Combined with adversarial noise $\textbf{r}_{adv}$ generated by Adaptive-VAT for perturbation, $\textbf{z}_{u}$ and $\textbf{z}_{l}$ are then fed into the decoder $D_{\theta}^{S}$ to produce final feature maps, which can be denoted as $D_{\theta}^{S}(\textbf{z}_{l} / \textbf{z}_{u} + \textbf{r}_{adv}) = \mathbb{R}_{L}^{S} / \mathbb{R}_{U}^{S}$.
The obtained feature map $\mathbb{R}$ corresponds to the marking points in the input image. Subsequently, we employ template matching to systematically verify the geometric relationships, determining whether these points align to form valid entrance lines of parking slots.

\textbf{Instructing-Teacher Branch.}
In this branch, the unlabeled data $\textbf{x}_{u}$ are processed through $E_{\theta}^{T}$ to produce the latent feature $\textbf{z}_{t}$, denoted as $E_{\theta}^{T}(\textbf{x}_{u}) = \textbf{z}_{t}$. The latent feature is then fed into the $D_{\theta}^{T}$ to generate final feature map $\mathbb{R}_{U}^{T}$, represented as $D_{\theta}^{T}(\textbf{z}_{t}) = \mathbb{R}_{U}^{T}$. During the training process, we employ adaptive moment estimation (Adam)~\cite{kingma2014adam} to minimize the loss function for $M_{\theta}^{S}$, concurrently updating $M_{\theta}^{T}$ using the  EMA~\cite{tarvainen2017mean} and parameters of $M_{\theta}^{S}$.

\subsection{Adaptive-VAT}
Feature perturbation is a prevalent strategy in self-supervised or semi-supervised learning. However, rather than adding random noise, we use VAT~\cite{miyato2018virtual} to estimate adversarial noise, which is designed to maximize the distance between the perturbed prediction and its original version. As shown in Fig. \ref{fig3}, given the detached latent feature $\textbf{z}$, the gradient-guided adversarial noise $\textbf{r}$ is calculated to maximize $(D_\theta(\textbf{z}+\textbf{r})-D_\theta(\textbf{z}) )$, and such noise is intended to confuse the network as much as possible. To achieve a satisfactory perturbation effect, the quality of $D_\theta$ is important as it determines the gradient direction. 

Liu et al. proposed T-VAT in PS-MT~\cite{liu2022perturbed}, setting $D_\theta$ to be $D_{\theta}^{T}$ for calculating the adversarial noise, which is not always optimal in our view. We observe that, in particular instances, especially during the early training stage, $D_{\theta}^{S}$ typically exhibits better quality, as it is guided by direct gradients. Therefore, we introduce Adaptive-VAT to apply the optimal disturbance. It adaptively selects the adversarial noise from $D_{\theta}^{T}$ or $D_{\theta}^{S}$ based on their resistance robustness in generating disturbances.
The adversarial noise $\textbf{r}_{adv}$ is computed based on the responses of $D_{\theta}^{T}$ and $D_{\theta}^{S}$ to ensure that the distance between the original and perturbed predictions is maximized, which should satisfy the following formula:
\begin{equation}               
\begin{aligned}
    \text{maximise }& \mathcal{M}\Big(d(D_{\theta}^{T}(\mathbf{z}_l / \mathbf{z}_u) , D_{\theta}^{T}(\mathbf{z}_l / \mathbf{z}_u + \textbf{r}_{adv})), \\
    &d(D_{\theta}^{S}(\mathbf{z}_l / \mathbf{z}_u), D_{\theta}^{S}(\mathbf{z}_l / \mathbf{z}_u + \textbf{r}_{adv}))
    \Big),\\
    \text{subject to }& ||\mathbf{r}_{adv}||_2 <=\epsilon, 
\end{aligned}
\end{equation}%
where $d(\cdot,\cdot)$ represents the mean squared error (MSE) loss, and $\mathcal{M}(\cdot,\cdot)$ selects the better decoder with lower MSE loss.

\subsection{Confidence-Guided Mask Consistency}
To construct the unsupervised constraint for unlabeled data, an intuitive idea is to employ MSE loss~\cite{tarvainen2017mean} between $\mathbb{R}_{U}^{S}$ and $\mathbb{R}_{U}^{T}$ to ensure the prediction consistency. 
However, this scheme might encourage potentially incorrect predictions to remain consistent, which could misguide the network and make it suboptimal to assign equal weight to all grid cells corresponding to $\mathbb{R}$.
To address this issue, as illustrated in Fig. \ref{fig3}, we introduce a confidence map that allocates attention unequally to different grid cells and provides a mask for regions with low confidence.

Specifically, for $\mathbb{R} \in \mathcal{S} \times \mathcal{S} \times \mathcal{N}$, we set $\mathcal{N} = 6+3$ and $\mathcal{S} = 16$ for $\textbf{x}$, meaning that we divide the input image $\textbf{x}$ into a $16 \times 16$ grid. The first 6 dimensions follow the setup of DMPR-PS~\cite{huang2019dmpr}, while the additional 3 dimensions capture the angle and classification details necessary for detecting slanted parking slots. 
In other words, the representation of a marking point consists of 9 components: $(x, y)$ for the reference point location, $(\cos\theta_{1/2}, \sin\theta_{1/2})$ for the angles of two edges, $\textbf{s}$ for the shape, $\textbf{t}$ for the type (\textit{i.e.}, perpendicular and slanted) of marking points, and $C$ for the confidence.

Therefore, as shown in Fig.\ref{fig3}, we develop the confidence-guided mask consistency loss $\mathcal{L}_{CGM}$ as:
\begin{equation}
\begin{cases}
    \sum_{i=1}^{\mathcal{S}^{2}} \{ ||C_{i}^{S} - C_{i}^{T}||_2 \} & \text{if } C_{i}^{T} < \tau \\
    \sum_{i=1}^{\mathcal{S}^{2}} \{ ||C_{i}^{S} - C_{i}^{T}||_2 + ||q_{i}^{S} - q_{i}^{T}||_2 \times C_{i}^{T} \} & \text{if } C_{i}^{T} \geq \tau
\end{cases}
,
\end{equation}%
where $q= (x,y,\cos\theta_1,\sin\theta_1,\cos\theta_2,\sin\theta_2,\textbf{s},\textbf{t})$, subscript $i$ denotes the cell index in the $16 \times 16$ grid, and $\tau$ is the confidence threshold (we set $\tau$ = 0.9). In this way, we not only screen out significant grid cells for supervision but also pay more attention to the grid cells that are more likely to contain marking points.

\subsection{Semi-Supervised Learning}
\textbf{Labeled Data.}
For labeled image pairs $(\textbf{x}_{l},\textbf{y}_{l})$, we can provide strong supervision for the prediction. Therefore, the supervised loss $\mathcal{L}_{sup}$ is defined as follows:
\begin{equation}
\begin{cases}
     \sum_{i=1}^{\mathcal{S}^{2}} \{ ||C_{i}^{S}||_2 \} & \text{if } \textbf{y}_{li} \ not \ exists \\
     \sum_{i=1}^{\mathcal{S}^{2}} \{ ||C_{i}^{S} - 1||_2 + ||q_{i}^{S} - \hat{q_{i}}||_2 \} & \text{if } \textbf{y}_{li} \ exists
\end{cases}
,
\end{equation}%
where $\hat{q_{i}}$ indicates the corresponding ground truth (GT) for the predictions, and $\textbf{y}_{li}$ exists only when the marking point is within the grid cell $i$.

\textbf{Unlabeled Data.}
For unlabeled data, we leverage the confidence-guided mask (CGM) consistency between $M_{\theta}^{T}$ and $M_{\theta}^{S}$ to establish an unsupervised constraint, as expressed in the equation below:
    \begin{equation}
    \mathcal{L}_{unsup}(\mathcal{D}_{U}, M_{\theta}^{S}, M_{\theta}^{T})=\mathcal{L}_{CGM}
\end{equation}%
The total objective function is composed of two terms based on the labeled and unlabeled data as follows:
\begin{equation}
\begin{array}{l}
    \mathcal{L}_{total}(\mathcal{D}_{L},\mathcal{D}_{U}, M_{\theta}^{S}, M_{\theta}^{T})= \\
    \mathcal{L}_{sup}(\mathcal{D}_{L}, M_{\theta}^{S})+ \beta \mathcal{L}_{unsup}(\mathcal{D}_{U}, M_{\theta}^{S}, M_{\theta}^{T})
\end{array}
,
\end{equation}%
where $\beta=\frac{N_{\mathcal{D}_{U}}}{N_{\mathcal{D}_{L}}}$ and $N_{\mathcal{D}_{U}}/N_{\mathcal{D}_{L}}$ represents the number of labeled/unlabeled samples.

\section{Experimental Results}
In this section, we first give the details of our experimental setup, and then compare the proposed SS-PSD with other solutions on different datasets using various partition protocols. Next, we present the ablation studies and detailed results across different real-world scenes. 

\subsection{Experimental Setup}
\textbf{Training.}
We used the Adam optimizer with an initial learning rate of $10^{-4}$ for training (batch size=24). The training was carried out on a single Nvidia 2080Ti GPU (12GB RAM) and an Intel(R) Xeon(R) CPU E5-2680 v4 @ 2.40GHz.

\textbf{Evaluation Metric.}
Precision and recall are sensitive to threshold selection, which can affect their reliability. To provide a more comprehensive assessment, we use average precision (AP) as our evaluation metric. For this evaluation, we apply the all-point interpolation method from the \textit{PASCAL VOC 2010}, as defined by the following equation:
\begin{equation}
    AP = \sum \Big((Recall_{j+1}-Recall_{j}) \times Precision_{j+1} \Big),\\
\end{equation}%
where $j$ is the index satisfying $Recall_{j} \neq Recall_{j+1}$.

The formula for computing recall and precision is as shown below:
\begin{equation}
    Precision =\frac{true\;positives}{true\;positives+false\;positives}
\end{equation}

\begin{equation}
    Recall =\frac{true\;positives}{true\;positives+false\;negatives} 
\end{equation}%
To determine the \textit{true positives}, we define a \textbf{marking point} as $P= \{ x, y, s, t, \theta_{1}, \theta_{2} \}$. Here, $(x, y)$ denotes the position of the marking point, $s$ indicates its shape, $t$ denotes its type, and $\theta_{1}, \theta_{2}$ represent the angles of the two edges in degrees. Let  $P^{g}=\{x^{g}, y^{g}, s^{g}, t^{g}, \theta_{1}^{g}, \theta_{2}^{g} \}$ be a GT marking point and $P^{d}=\{x^{d}, y^{d}, s^{d}, t^{d}, \theta_{1}^{d}, \theta_{2}^{d} \}$ be a detected marking point. We define the following conditions:
\begin{gather}
    (x^{g}-x^{d})^{2} + (y^{g}-y^{d})^{2} < I^{2} \\
    |\theta_{1}^{g} - \theta_{1}^{d}| < B \ or \ 360^{\circ} - |\theta_{1}^{g} - \theta_{1}^{d}| < B \\
    |\theta_{2}^{g} - \theta_{2}^{d}| < B \ or \ 360^{\circ} - |\theta_{2}^{g} - \theta_{2}^{d}| < B \\
    s^{g} = s^{d} \ and \ t^{g} = t^{d}
\end{gather}%
If these conditions are met, we consider $P^{g}$ to be accurately identified and $P^{d}$ to be a \textit{true positive}.

We define the \textbf{parking slot} as $S=\{(x_{1}, y_{1}), (x_{2}, y_{2}), \theta_{s}\}$, where $(x_{1}, y_{1})$ and $(x_{2}, y_{2})$ are the coordinates of two marking points, and $\theta_{s}$ denotes the direction of the entrance line. For a GT parking slot $S_{g}=\{(x_{1}^{g}, y_{1}^{g}), (x_{2}^{g}, y_{2}^{g}), \theta_{s}^{g}\}$, a detected parking slot $S_{d}=\{(x_{1}^{d}, y_{1}^{d}), (x_{2}^{d}, y_{2}^{d}), \theta_{s}^{d}\}$ is considered a \textit{true positive} if it satisfies the following equation: \begin{gather}
    (x_{1}^{g}-x_{1}^{d})^{2} + (y_{1}^{g}-y_{1}^{d})^{2} < I^{2}  \\
    (x_{2}^{g}-x_{2}^{d})^{2} + (y_{2}^{g}-y_{2}^{d})^{2} < I^{2} \\
    |\theta_{s}^{g} - \theta_{s}^{d}| < B \ or \ 360^{\circ} - |\theta_{s}^{g} - \theta_{s}^{d}| < B
\end{gather}%
A parking slot is considered a \textit{true positive} if its two points are within $I$ pixels of the GT and the angle difference between $S_{d}$ and $S_{g}$ is less than $B$. For ps2.0, $I=10$; for CRPS-D, $I=8.53$, adjusted from $10 \times \frac{512}{600}$  to account for the smaller image size of $512 \times 512$.

\begin{table*}[!t]
\centering
\caption{Comparison with SoTA approaches using few-supervision with different ratios of labeled data on the proposed dataset CRPS-D.}
\begin{tabular}{lcccccc}
\toprule
Ratio & 1/24 & 1/12 & 1/8  & 1/4  & 1/2  & 1/1 \\
\midrule
   & 1242 labels & 2484 labels & 3726 labels & 7452 labels & 14901 labels & 29803 labels \\
   & 29803 images & 29803 images & 29803 images & 29803 images & 29803 images & 29803 images \\
\midrule
DMPR-PS$^{*}$\cite{huang2019dmpr}  & 52.70\% & 66.84\% & 70.28\% & 76.98\% & 83.23\% & 87.20\% \\
GCN$^{*}$\cite{min2021attentional}   & 55.23\% & 63.32\% & 67.80\% & 72.57\% & 76.31\% & 79.44\% \\
\textbf{Ours} & \textbf{71.72\%} & \textbf{79.75\%} & \textbf{81.03\%} & \textbf{83.78\%}  & \textbf{86.51\%} & \textbf{/} \\
\bottomrule
\end{tabular}
\label{tab5}
\end{table*}

\begin{table*}[!t]
\centering
\caption{Comparison with SoTA approaches using few-supervision with different ratios of labeled data on ps2.0.}
\begin{tabular}{lcccccc}
\toprule
Ratio & 1/24 & 1/12 & 1/8  & 1/4  & 1/2  & 1/1 \\
\midrule
   & 410 labels & 819 labels & 1230 labels & 2457 labels & 4913 labels & 9827 labels \\
   & 9827 images & 9827 images & 9827 images & 9827 images & 9827 images & 9827 images \\
\midrule
    DMPR-PS$^{*}$\cite{huang2019dmpr}  & 60.95\% & 78.05\% & 87.28\% & 93.70\% & 96.47\% & 98.64\% \\
    GCN$^{*}$\cite{min2021attentional}  & 76.70\% & 85.72\% & 92.78\% & 95.47\% & 97.45\% & 98.31\% \\
    \textbf{Ours} & \textbf{80.44\%} & \textbf{90.47\%} & \textbf{94.52\%}  & \textbf{96.10\%} & \textbf{97.79\%} & \textbf{/} \\
\bottomrule
\end{tabular}
\label{tab6}
\end{table*}

\subsection{Results on Different Partition Protocols}
Consistent with the Mean Teacher \cite{tarvainen2017mean}, we evaluate our approach by sub-sampling datasets with a ratio of $1/n$ for the labeled set and $(1 - \frac{1}{n})$ for the unlabeled set. 
We use labeled ratios of 1/24, 1/12, 1/8, 1/4, and 1/2 for both the proposed CRPS-D dataset and the publicly available dataset ps2.0.
As our SS-PSD is the first semi-supervised solution in this field, we can only compare our method with the fully supervised solutions that are publicly available.

\textbf{CRPS-D.}
As shown in Table~\ref{tab5}, with a 1/24 ratio (1242 labeled images), our method achieves improvements of 19.02\% and 16.49\% over DMPR-PS~\cite{huang2019dmpr} and GCN~\cite{min2021attentional} on $AP_{parking-slot}$, respectively. Even across other ratios (1/12, 1/8, and 1/4), our approach consistently enhances performance by 6\% to 12\%. These results demonstrate that our method can make good use of the unlabeled data, leading to significant gains. The $^{*}$ indicates the approaches that have undergone retraining.

\textbf{ps2.0.}
We also tested on the ps2.0 dataset to demonstrate the generalization capability of our approach across different parking slot datasets. Table~\ref{tab6} reveals that our model surpasses supervised baselines across all partition protocols.

\begin{figure}[ht]
    \centering
	  \subfloat[]{
       \includegraphics[width=0.49\linewidth]{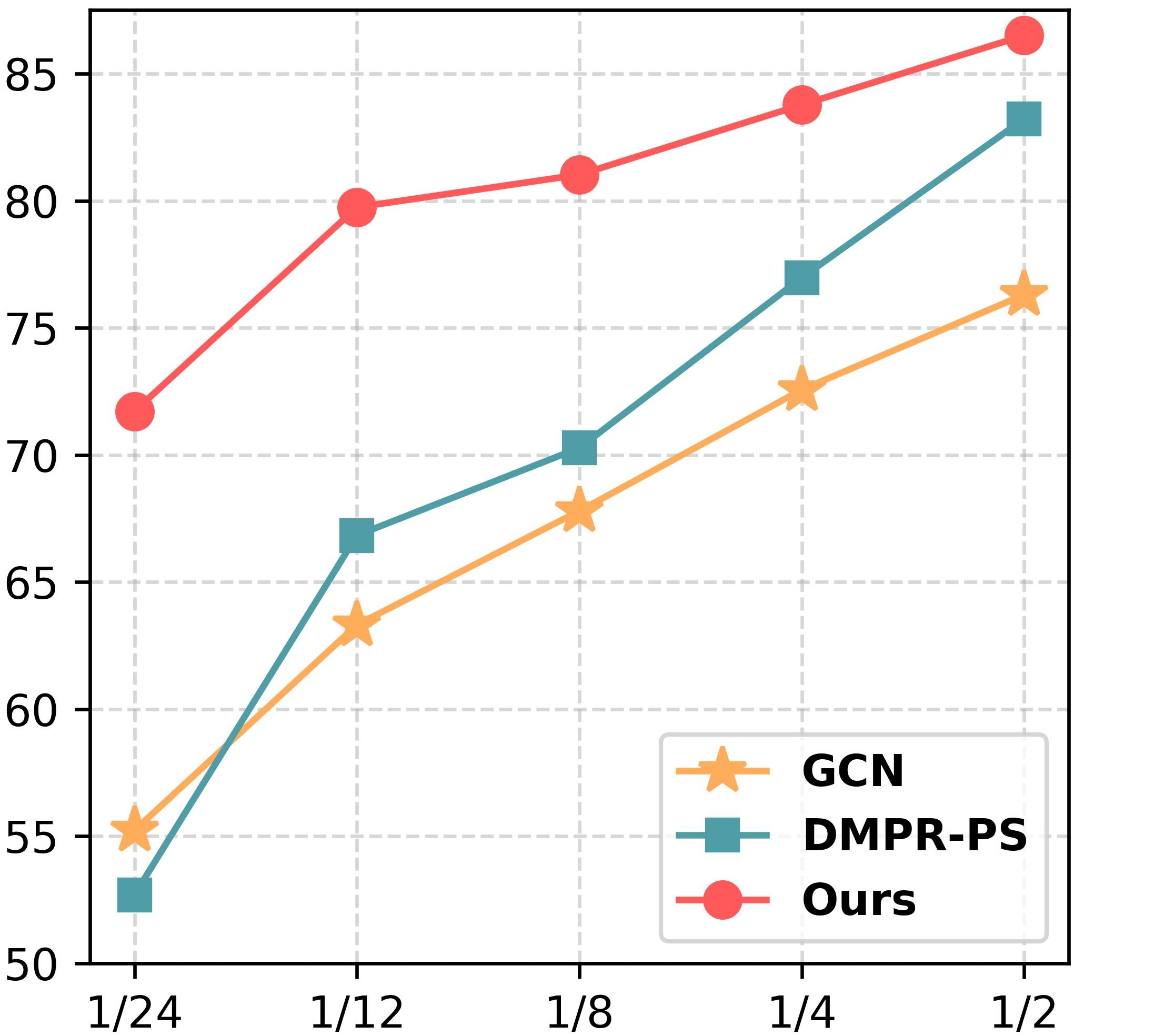}}
    \label{5a}
	  \subfloat[]{
        \includegraphics[width=0.46\linewidth]{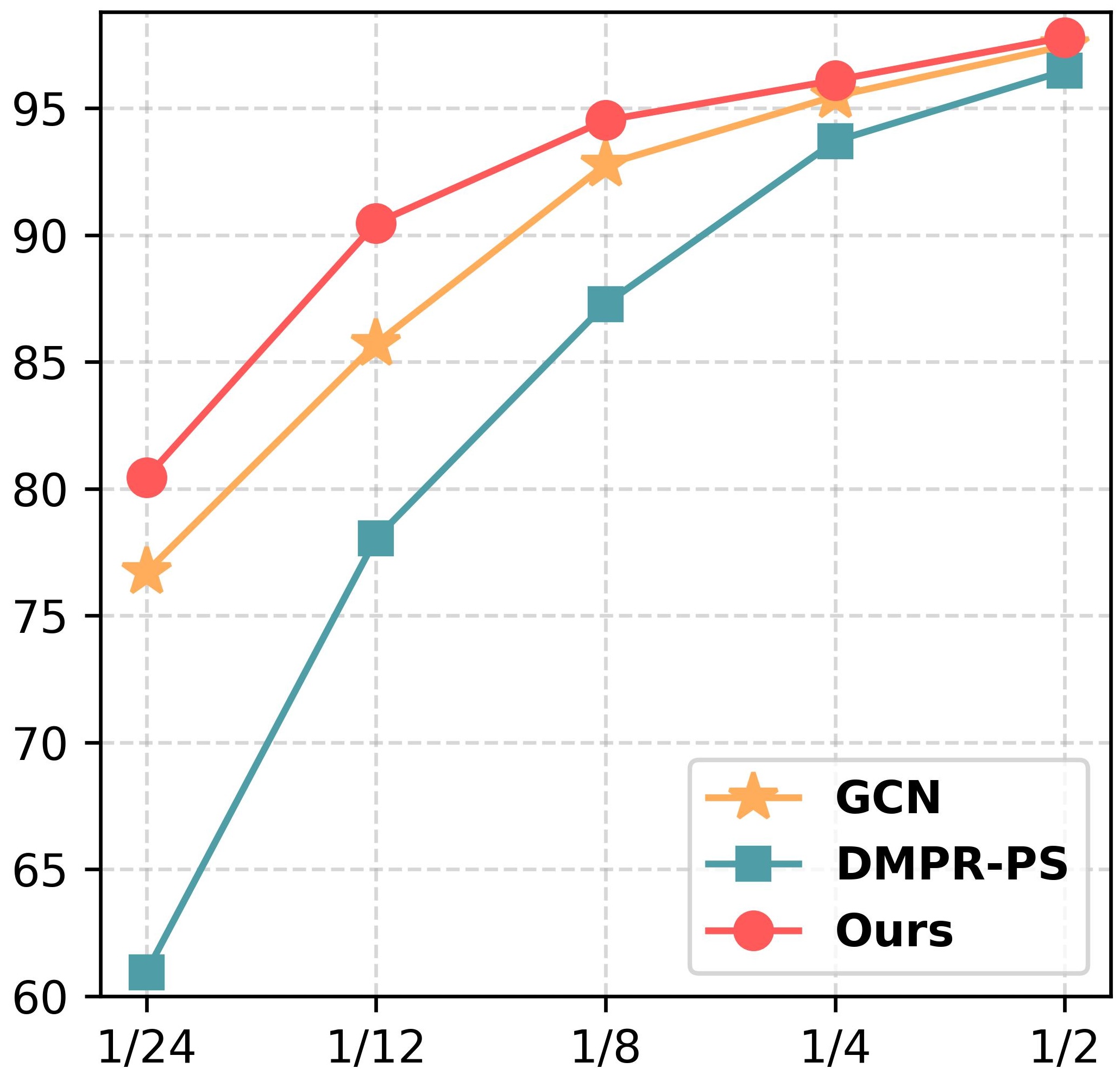}}
    \label{5b}
	  \caption{\textbf{Improvements over the supervised baseline.} Results of our approach and the supervised baseline on (a) CRPS-D and (b) ps2.0 with different partition protocols.}
    \label{fig_curve} 
\end{figure}

Fig. \ref{fig_curve} illustrates that our model shows a more pronounced improvement on the CRPS-D dataset over the supervised baselines, particularly with smaller labeled data, due to the greater complexity and realism of the CRPS-D dataset relative to the ps2.0 dataset. The proposed semi-supervised approach effectively captures a broader range of features from the CRPS-D dataset.

\begin{table*}[!t]
\centering
\caption{Ablation study using the 1/12 labeled ratio on CRPS-D.}
\begin{tabular}{cccccc}
\toprule
   Supervised Baseline & MT (with MSE Loss) & CGM Consistency  & Adaptive-VAT & $AP_{marking-point}$ & $AP_{parking-slot}$ \\
\midrule
    \checkmark  &  &  &  & 72.47\% & 66.84\% \\
     \checkmark & \checkmark &  &  & 78.77\% & 75.33\% \\
     \checkmark & \checkmark & \checkmark &  & 80.86\% & 78.93\% \\
     \checkmark & \checkmark & \checkmark & \checkmark & \textbf{81.59\%} & \textbf{79.75\%} \\
\bottomrule
\end{tabular}
\label{tab7}
\end{table*}

\subsection{Effective Use of Unlabeled Data}
We would like to clarify that our work introduces the \textbf{first semi-supervised} approach for parking slot detection, so directly comparing it with fully supervised methods is not completely fair.

Despite that, we are willing to provide more comparative results with recent methods. Note, these methods (\textit{e.g.}, SCCN(2020)\cite{zheng2022center}, Bui et al.(2023)\cite{bui2023one} and SAM-PS(2024)\cite{zhai2024sam}) are not open-source, so we can only report the performance on ps2.0 dataset from their original papers. The results are shown in Table~\ref{tab_recent}, where ‘Ours (1/n)’ indicates the use of 1/n labels for training.
When using only 1/12 of the labels (819), the performance of our semi-supervised method is significantly better than that of existing self-supervised methods (SAM-PS\cite{zhai2024sam}). Furthermore, when we use half of the labels (4913), we achieve performance comparable to the fully supervised model (\textit{e.g.}, VPS\cite{li2020vacant}), with only about a 1\% difference.

\begin{table}[ht]
\centering
\setlength{\tabcolsep}{1.8mm}
\caption{Comparison results with recent methods.}
\begin{tabular}{lccccc}
\toprule
    Method & Labels & Images &  Precision(\%) & Recall(\%) & F1(\%) \\
\midrule
    SAM-PS\cite{zhai2024sam} & 0 & 9827 & 94.29 & 81.35 & 87.34 \\
    \textbf{Ours (1/12)} & \textbf{819} & 9827 & \textbf{95.45} & \textbf{91.30} & \textbf{93.34} \\
\midrule
       DeepPS\cite{zhang2018vision}  & 9827 & 9827 & 98.99 & 99.13 & 99.06 \\
       DMPR-PS\cite{huang2019dmpr} & 9827 & 9827 & 99.42 & 99.37 & 99.39 \\
       VPS\cite{li2020vacant}  & 9827 & 9827 & 99.63 & 99.10 & 99.36 \\
       GCN\cite{min2021attentional}  & 9827 & 9827 & 99.56 & 99.42 & 99.49\\
       SCCN \cite{zheng2022center}  & 9827 & 9827 & 99.35 & 99.17 & 99.26 \\
       Bui et al.\cite{bui2023one} & 9827 & 9827 &  99.63 & 99.63 & 99.63 \\
       \textbf{Ours (1/2)} & \textbf{4913}  & 9827 & 98.64 & 98.07 & 98.35\\
\bottomrule
\end{tabular}
\label{tab_recent}
\end{table}

\begin{figure}[htbp]
    \centering
    \includegraphics[width=1\linewidth]{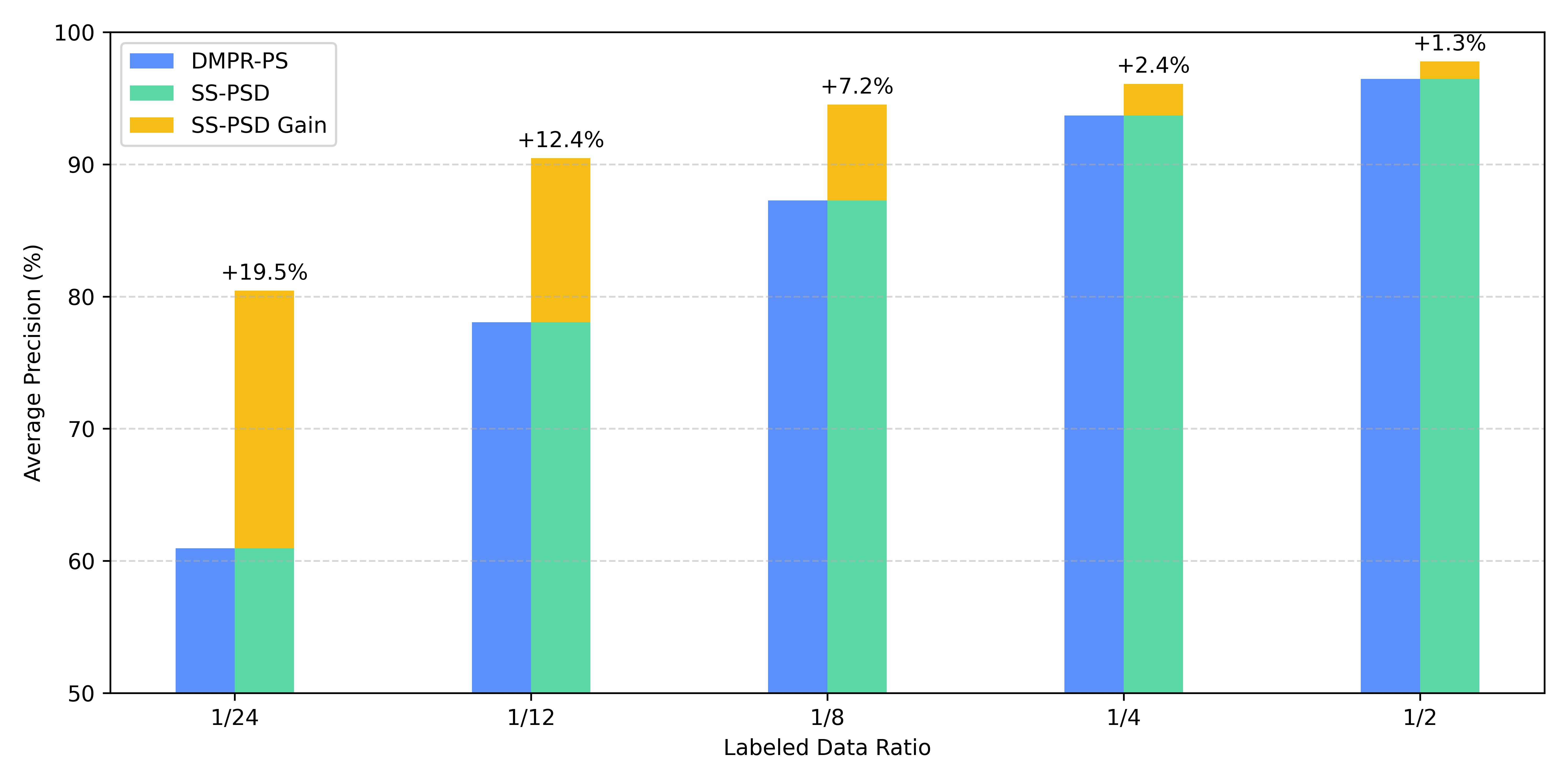}
    \caption{\textbf{Performance comparison} between the proposed SS-PSD and the supervised baseline under varying labeled ratios on the \textbf{ps2.0 dataset}. The yellow segments indicate the performance improvement brought by SS-PSD.}
    \label{fig:ps}
   \end{figure}

\begin{figure}[htbp]
    \centering
    \includegraphics[width=1\linewidth]{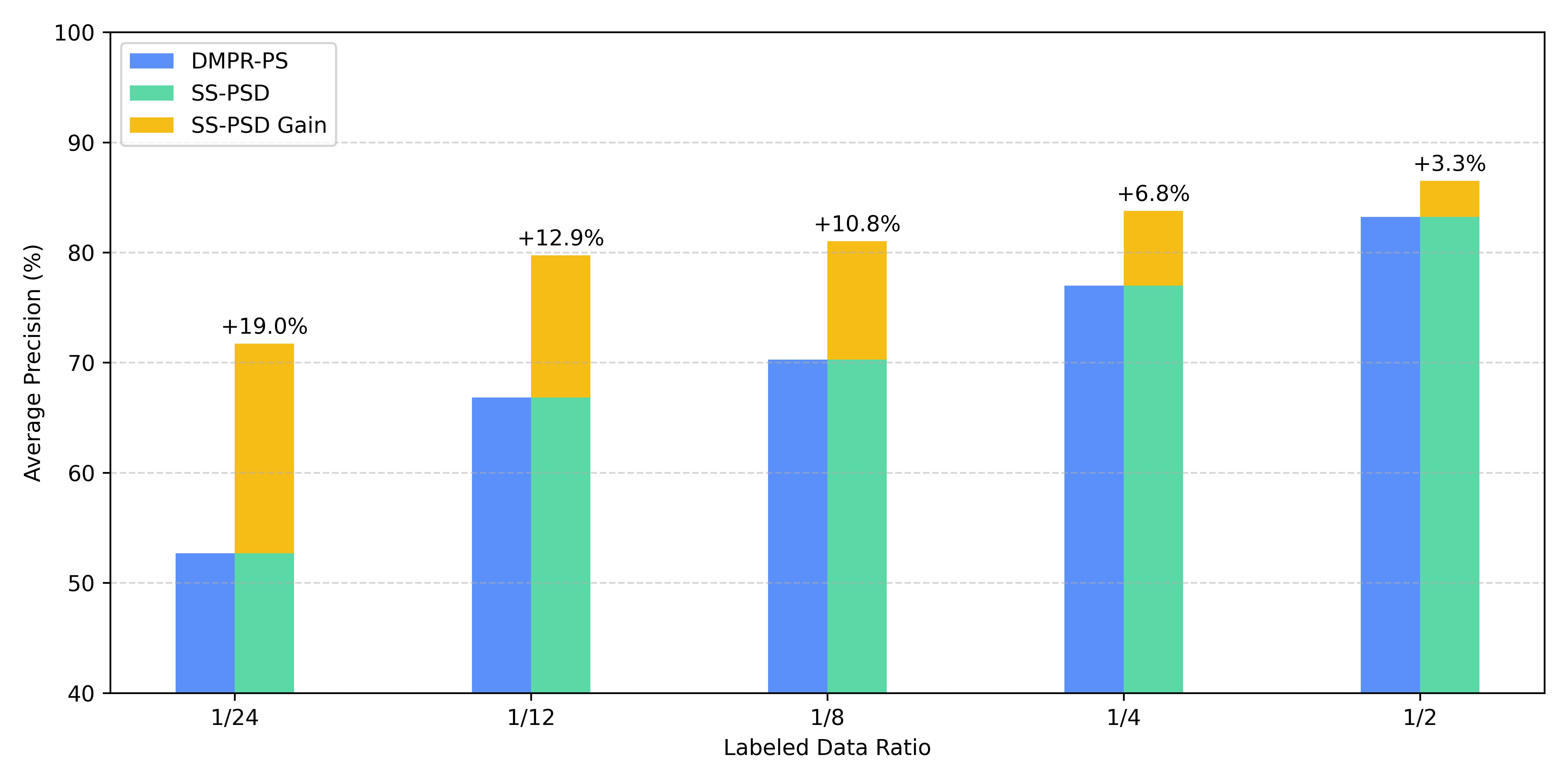}
    \caption{\textbf{Performance comparison} between the proposed SS-PSD and the supervised baseline under varying labeled ratios on \textbf{CRPS-D dataset}. The yellow segments indicate the performance improvement brought by SS-PSD.}
    \label{fig:crps}
   \end{figure}

In addition, we illustrate the performance \textbf{gap} between SS-PSD and the fully supervised model (our baseline, DMPR-PS \cite{huang2019dmpr}) as the proportion of labeled data increases. As shown in Fig.\ref{fig:ps} and Fig.\ref{fig:crps}, SS-PSD consistently outperforms the fully supervised baseline when only a small fraction of labeled data is available. As the amount of labeled data increases, this performance gap gradually narrows. Such diminishing returns are expected, as the marginal benefit of leveraging unlabeled data decreases once sufficient labeled data is available.

These results highlight the effectiveness of our method in label-scarce scenarios and its scalability as supervision increases, making it a practical solution for real-world applications where annotation resources are limited.

\subsection{Ablation Study}
Here, we explore the impact of the confidence-guided mask (CGM) consistency and the Adaptive-VAT perturbation on our method. All experiments are conducted on the CRPS-D dataset with a labeled data ratio of 1/12. The contributions of each component are outlined in Table~\ref{tab7}. We present the detection results for both marking points and parking slots concurrently. Compared with the semi-supervised baseline MT~\cite{tarvainen2017mean} with MSE loss, CGM consistency improves the $AP_{parking-slot}$ by 3.60\%, while Adaptive-VAT can further provide an additional 0.82\% improvement.


Additionally, we performed ablation experiments on CGM consistency and Adaptive-VAT separately to further demonstrate their effectiveness.

\textbf{CGM Consistency.}
We delved into the effects of various components on CGM consistency.
Initially, we removed the mask to derive CG consistency, followed by the removal of confidence weights to achieve C consistency.
Table~\ref{tab9} illustrates that equally constraining all cells to be consistent cannot lead to satisfactory results (55.48\% $AP_{parking-slot}$). 
However, leveraging confidence to guide the weight allocation can notably enhance the results, while masking low-confidence areas can further boost the outcomes by 0.71\%. 

\begin{table}[ht]
\centering
\caption{Ablation study of various components on CGM Consistency and Adaptive-VAT. }
\setlength{\tabcolsep}{4.0mm}
\begin{tabular}{lcc}
\toprule
  Methods &  $AP_{marking-point}$ & $AP_{parking-slot}$ \\
\midrule
      C consistency &   60.07\% & 54.88\% \\
      CG consistency &   80.75\% & 78.22\% \\
      CGM consistency &  \textbf{81.86\%} & \textbf{78.93\%} \\
\midrule
        S-VAT &   80.97\% & 78.87\% \\
       T-VAT &   81.16\% & 79.53\% \\
        Adaptive-VAT &  \textbf{81.59\%} & \textbf{79.75\%} \\
\bottomrule
\end{tabular}
\label{tab9}
\end{table}

\textbf{Adaptive-VAT.}
Removing the adaptability of Adaptive-VAT leads to its degradation to `S-VAT'~\cite{tarvainen2017mean} or T-VAT~\cite{liu2022perturbed}.
As shown in Table~\ref{tab9}, applying Adaptive-VAT results in better performance, as it adaptively selects the better strategy between `S-VAT' and T-VAT.

\begin{table}[ht]
\centering
\caption{Results of different confidence thresholds trained using a 1/12 labeled ratio on CRPS-D.}
    \begin{tabular}{cccc}
    \toprule
    Threshold ($\tau$) & $AP_{marking-point}$ & $AP_{parking-slot}$ & Epochs\\
    \midrule
    0.1 & 81.12\% & 78.81\%	& 36 \\
    0.2	& 81.59\%	& 79.47\%	& 22 \\
    0.3	& 81.28\%	& 79.08\%	& 21 \\
    0.4	& 81.13\%	& 79.34\%	& 23 \\
    0.5	& 81.06\%	& 79.01\%	& 17 \\
    0.6 & 80.75\%	& 79.10\%	& 40 \\
    0.7	& 81.38\%	& 79.50\%	& 23 \\
    0.8	& 80.75\%	& 78.95\%	& 26 \\
    0.9	& \textbf{81.59\%}	& \textbf{79.75\%}	& 22 \\
    \bottomrule
    \end{tabular}
\label{tab_tau}
\end{table} 

\subsection{Selection of Hyper-Parameter}
This section discusses the selection of values of two key hyper-parameters: the confidence threshold ($\tau$) and the disturbance intensity ($eps$).

\textbf{Confidence Threshold ($\tau$).} 
Following FixMatch~\cite{sohn2020fixmatch}, which improves pseudo-label accuracy by selecting labels with higher confidence, we set a higher threshold ($\tau$ = 0.9). However, we also experimented with confidence threshold values ranging from 0 to 1, in increments of 0.1. As shown in Table~\ref{tab_tau}, a threshold of 0.9 produced the best performance.

\textbf{Disturbance Intensity ($eps$).}
To demonstrate that our Adaptive-VAT achieves superior performance across various disturbance intensities, we varied the disturbance levels from high to low, using the values of $eps$  = 10, 1, and 0.1.
As shown in Table~\ref{tab_eps}, applying our Adaptive-VAT results in better performance compared with S-VAT and T-VAT. Based on the experimental results, we have ultimately selected $eps = 0.1$.

\begin{table}[ht]
\centering
\caption{Results of different disturbance intensities trained using a 1/12 labeled ratio on CRPS-D.}
\begin{tabular}{ccccccc}
\toprule
 & \multicolumn{3}{@{}c@{}}{$AP_{marking-point}$} & \multicolumn{3}{@{}c@{}}{$AP_{parking-slot}$} \\\cmidrule{2-4}\cmidrule{5-7}%
   \multirow{-2}{*}{$eps$}  & S-VAT & T-VAT & Ours & S-VAT & T-VAT & Ours \\
\midrule
    10.0 & 80.89\% & 80.78\% & 81.32\% &79.11\% & 78.89\% & 79.15\%  \\
     1.0 & 80.87\% & 81.04\% & 81.68\% &78.65\% &78.89\% & 79.54\%  \\
     0.1 & 80.97\% & 81.16\% & \textbf{81.59\%} & 78.87\% & 79.53\% & \textbf{79.75\%}  \\
\bottomrule
\end{tabular}
\label{tab_eps}
\end{table}

\begin{figure*}[!t]
\centering
\includegraphics[width=0.95\textwidth]{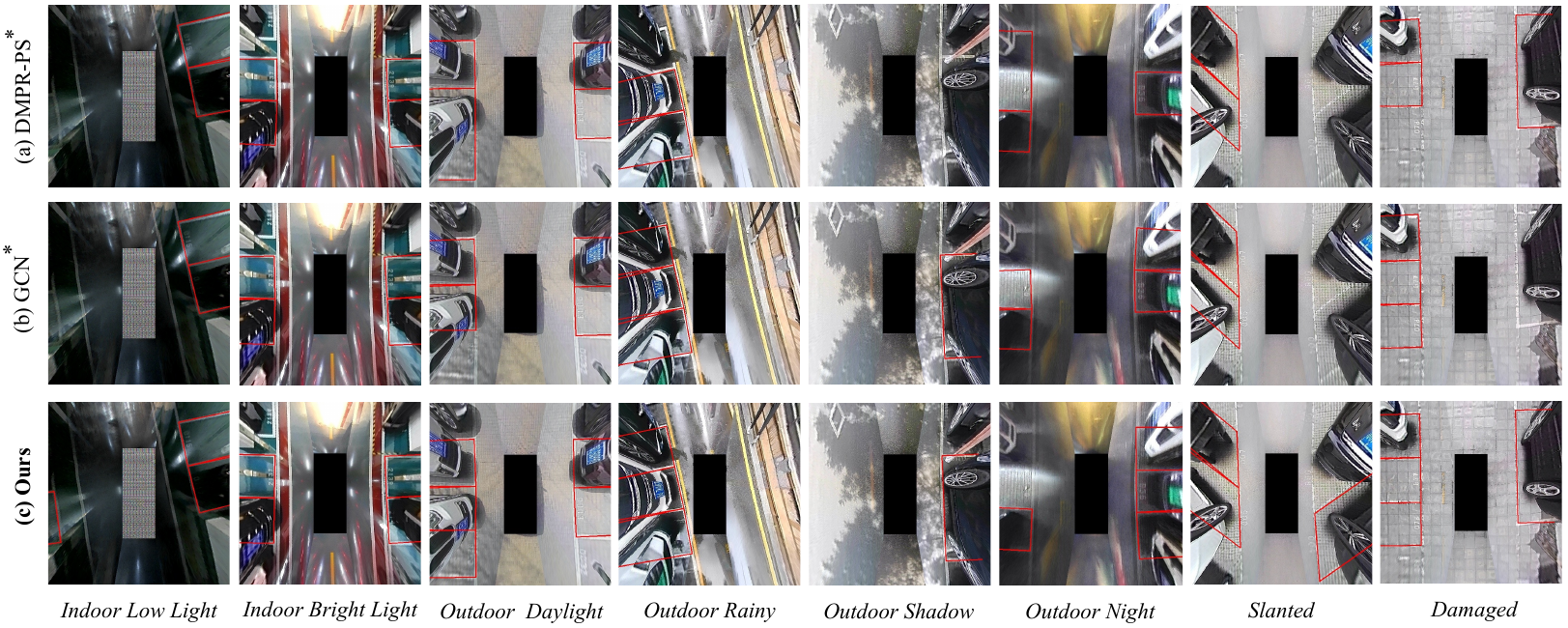}
\caption{\textbf{Visualized results of different scenes.} Results of (a) DMPR-PS, (b) GCN  and (c) our SS-PSD trained on the proposed CRPS-D dataset with 1/12 labeled data. We apply this dataset in automated parking systems; refer to our supplementary video.}
\label{fig6}
\end{figure*}

    \begin{figure*}[!t]
    \centering
    \includegraphics[width=0.97\linewidth]{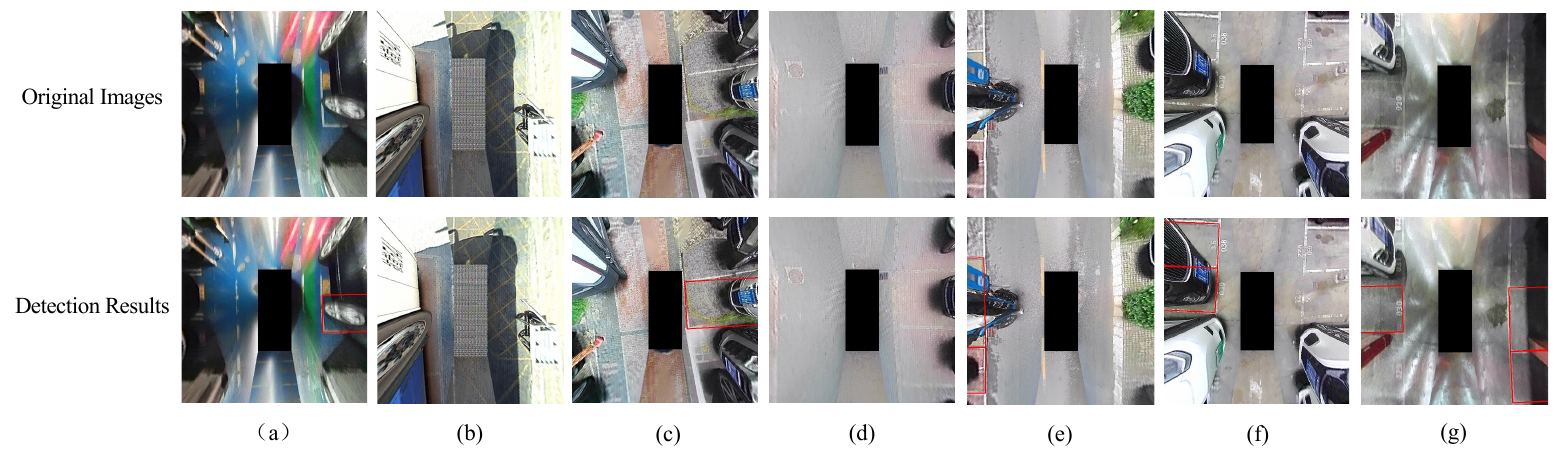}
    \caption{
    \textbf{Representative failure cases of SS-PSD in real-world parking scenarios.} Each column (a–g) illustrates a challenging condition: (a) complex lighting in an underground garage, (b) outdoor glare, (c) heavy rain, (d) faded markings, (e) severe occlusion, (f) damaged slot lines, and (g) nighttime scenes with low illumination. The top row shows the input bird’s-eye view images, while the bottom row presents the corresponding detection results.
    }
    \label{fig:fail}
   \end{figure*}

\subsection{Evaluation Results on Diverse Scenes}
Table~\ref{tab8} shows the performance evaluations of the supervised baseline (DMPR-PS\cite{huang2019dmpr}) and our SS-PSD approach across different scenes (corresponding to sub-test sets of the CRPS-D dataset). The results are trained on the CRPS-D dataset with a 1/12 ratio of labeled data and 11/12 unlabeled data. Our model performs better in both simple conditions (\textit{e.g.}, `indoor low light', `outdoor shadow') and challenging (\textit{e.g.}, `damaged', `outdoor night') scenarios. The visualized results of parking slot detection are illustrated in Fig. \ref{fig6}. 
In particular, our semi-supervised solution can significantly alleviate the troublesome issue of \textbf{long-tailed distribution} by explicitly exploiting the easily available unlabeled data.
For instance, the ‘outdoor night' subset has the smallest data scale and shows the worst performance in the supervised scheme, while our SS-PSD could greatly improve the model performance.

\begin{table}[ht]
\centering
\caption{Evaluation results on the sub-test sets of CRPS-D.}
\setlength{\tabcolsep}{4.0mm}
\begin{tabular}{lccc}
\toprule
 \multicolumn{2}{c}{} & \multicolumn{2}{c}{$AP_{parking-slot}$} \\ \cmidrule{3-4}%
   \multirow{-2}{*}{\textit{Subset}} & \multirow{-2}{*}{$N_{images}$} & DMPR-PS$^{*}$ & \textbf{Ours} \\
\midrule
      \textit{Indoor Low Light} & 545  & 74.15\% & 86.68\%  \\
      \textit{Indoor Bright Light} & 786  & 70.00\% & 84.80\% \\
      \textit{Outdoor Shadow} & 1389  & 69.19\% & 81.09\%\\
      \textit{Outdoor Daylight} & 1703  & 68.06\% & 78.99\% \\
\midrule
      \textit{Slanted} & 802  & 61.61\% & 76.79\% \\
      \textit{Damaged} & 302  & 58.90\% & 71.83\% \\
      \textit{Outdoor Rainy} & 323  & 56.07\% & 71.07\% \\
      \textit{Outdoor Night} & 86  & 38.38\% & 63.45\% \\
\bottomrule
\end{tabular}
 \label{tab8}
\end{table}

\subsection{Analysis of Failure Cases}

Although the SS-PSD model demonstrates impressive performance in most scenarios, it does have limitations when confronted with extreme environmental conditions. In this section, we will analyze several failure cases. Several representative failure cases in real-world parking scenarios are presented in Fig. \ref{fig:fail}, highlighting the typical challenges that impact the model’s performance:

\textbf{Column (a).} 
Column (a) shows the effect of complex lighting conditions in an underground garage, where poor illumination, shadows, and floor reflections make slot detection difficult. These reflections can be mistakenly interpreted as parking slot lines, leading to false positives.

\textbf{Column (b).} 
In column (b), strong reflections caused by outdoor glare further impair the model’s ability to detect parking slots. 

\textbf{Column (c).}
Column (c) illustrates how heavy rain degrades detection accuracy due to reduced contrast, surface reflections, and partially obscured parking slot markings.

\textbf{Column (d).}
Column (d) shows that faded markings make it difficult for the model to discern slot boundaries, especially when the lines are significantly worn.

\textbf{Column (e).}
Column (e) depicts cases of severe occlusion, where vehicles or other objects block the view of parking slots.

\textbf{Column (f).}
Column (f) shows how heavily damaged slot lines, such as those that are faded or partially erased, can result in missed detections.

\textbf{Column (g).} 
Finally, column (g) shows nighttime scenarios, where limited lighting and sensor noise significantly degrade visual clarity. Markings are often barely visible in poorly lit areas, resulting in missed detections.

\begin{table}[ht]
\centering
\caption{Detection performance across scene types under common failure conditions.}
\setlength{\tabcolsep}{2.5mm}
    \begin{tabular}{llcc}
        \toprule
         Factor   & Scene Type & $AP_{parking-slot}$ & $\Delta AP$ 
 \\
        \midrule
             \multirow{4}{*}{Lighting} & \textit{Indoor Low Light} & 86.68\% & –  \\
             & \textit{Indoor Bright Light} & 84.80\% & ↓1.88\%  \\
              & \textit{Outdoor Daylight} & 78.99\% & ↓7.69\%   \\
              & \textit{Outdoor Night} & 63.45\% & ↓23.23\%  \\
        \midrule
          Shadow & \textit{Outdoor Shadow} & 81.09\% & ↓5.59\% \\
        \midrule
          Rain & \textit{Outdoor Rainy} & 71.07\% & ↓15.61\% \\
        \midrule
          Damage &  \textit{Damaged} & 71.83\% & ↓14.85\% \\
        \bottomrule
    \end{tabular}
    \label{tab_fail}
\end{table}

These failure cases reveal the model’s limitations under adverse conditions such as poor visibility and occlusion. Despite these challenges, SS-PSD generally outperforms fully supervised methods in scenarios involving moderate occlusion or mild degradation. Future work may improve robustness by incorporating additional data sources, such as temporal cues from video sequences or sensor fusion.

To further substantiate the qualitative observations, we conducted a quantitative analysis to assess how specific environmental factors influence detection performance. As shown in Table~\ref{tab_fail}, we group scenes by key types of visual degradation, including lighting conditions, shadows, rainfall, and marking damage, and report the corresponding \textit{AP} scores. Using the `Indoor Low Light' scenario as a reference, we observe notable performance drops under challenging conditions such as Outdoor Night (23.23\%), Outdoor Rainy (15.61\%), and Damaged markings (14.85\%). These results provide quantitative support for the failure cases and offer deeper insight into the model’s limitations in complex real-world environments.

\subsection{Application in Real-World Scenario}
\begin{figure}[!t]
\centering
\includegraphics[width=0.8\linewidth]{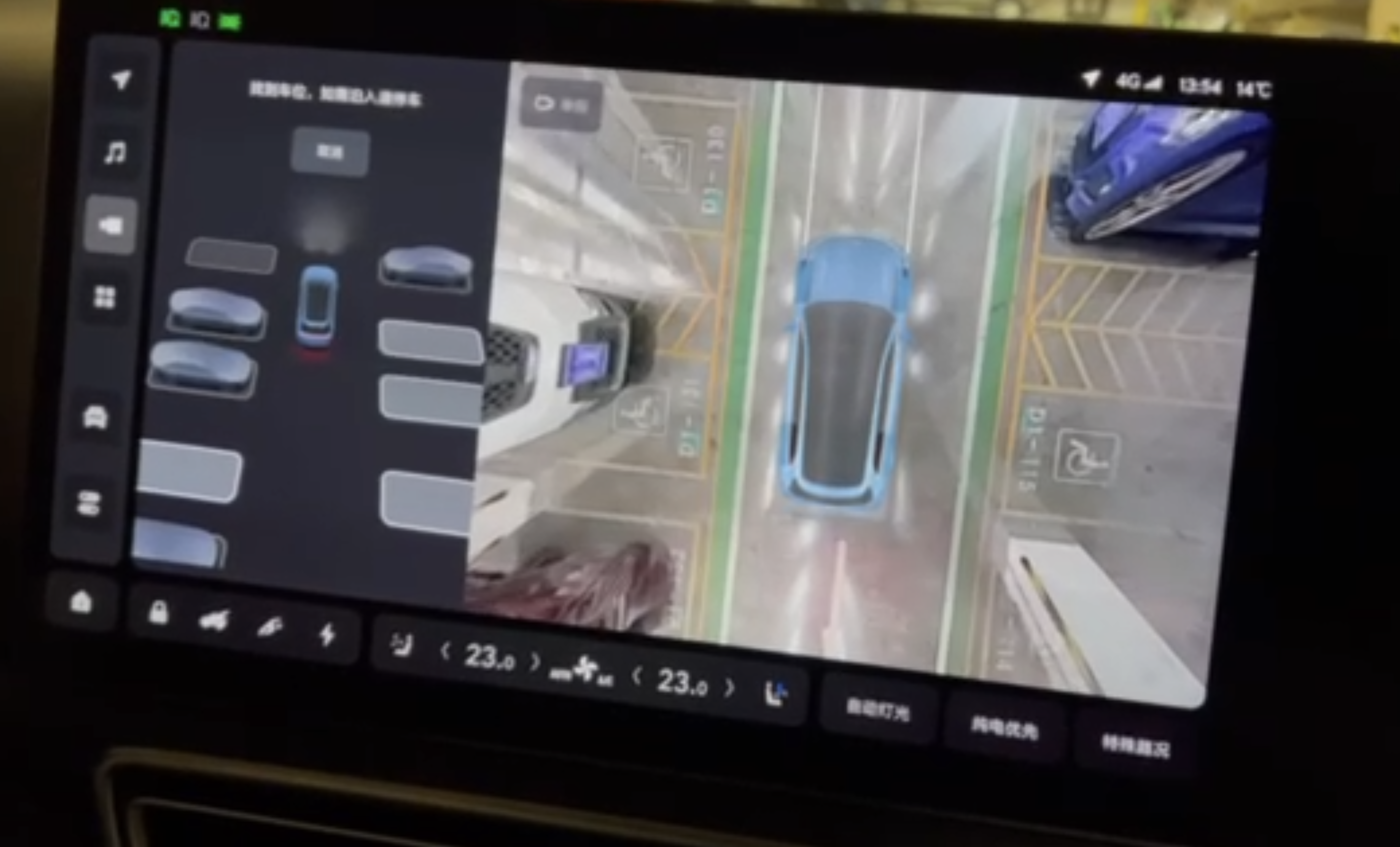}
\caption{We have developed an \textbf{ intelligent parking system} and integrated the model \textbf{trained on the CRPS-D dataset} into it. The system can automatically guide vehicles into parking slots after detecting an available slot.}
\label{fig_e1}
\end{figure}

\begin{figure}[!t]
\centering
\includegraphics[width=0.5\linewidth]{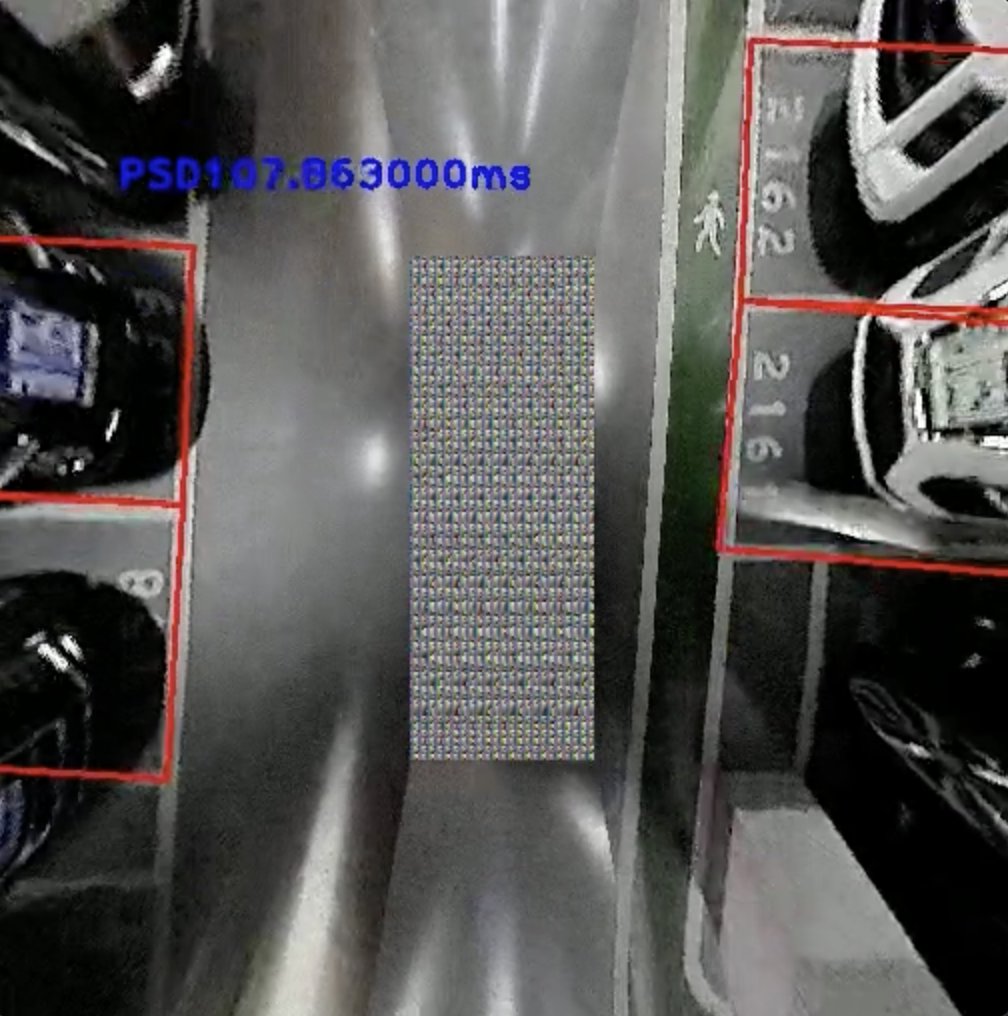}
\caption{Visualization of the entire parking slot detection process via screen recording. See supplementary video for details.}
\label{fig_test}
\end{figure}

We hope this dataset will contribute to the advancement of real-world parking slot detection. To demonstrate the practical applicability of our technique, we trained our model on the CRPS-D dataset and tested it in real-world road environments, as shown in Fig. \ref{fig_e1}. The test was conducted in an indoor parking lot with relatively poor lighting conditions, which presents a challenging scenario for typical detection systems. As shown in Fig. \ref{fig_test}, our model produced satisfactory results, accurately detecting parking slots even under these suboptimal conditions. Additionally, we have included a video, titled{`\textbf{multimedia\_appendix.mp4}'}, in the supplementary material, which demonstrates the model’s real-time performance. 

In addition, we deployed a trained SS-PSD model on an automotive-grade processor equipped with a Neural Processing Unit (NPU), achieving an inference speed of \textbf{20 FPS}, which meets the latency requirements for low-speed automated parking systems. 
The two lightweight modules introduced in our framework, namely CGM Consistency and Adaptive-VAT, are only active during training and do not impact the model architecture or inference cost.
While semi-supervised learning naturally incurs a higher training overhead due to the use of unlabeled data and consistency regularization, this is a common and acceptable trade-off, given the reduced annotation demand and the performance gains achieved.

In the future, we plan to optimize the model further to handle a wider range of environmental conditions and expand its functionality to support other traffic-related applications. We also aim to release this model to the public in the near future to encourage broader adoption and further research in this area.

\section{Conclusion}
We have developed a large-scale benchmark dataset for real-world automated parking scenarios, consisting of bird’s-eye view images captured under diverse weather conditions and across various scenes. This dataset poses greater challenges compared to previous ones, featuring a larger size, a wider variety of scenarios, a higher proportion of slanted parking slots, and more challenging instances with damaged lines. Additionally, we introduce a semi-supervised baseline that employs a teacher-student model with confidence-guided mask consistency and adaptive feature perturbation. Our approach achieves SoTA performance on both publicly available datasets and our newly introduced dataset, especially in the absence of labeled data.

Furthermore, although the current work focuses on detecting parking slot markings, we plan to extend our approach to include occupancy detection in future research. By incorporating a classification-based module, we aim to determine whether each detected slot is occupied or vacant, thereby enabling a more comprehensive and practical automated parking system.

We also plan to explore the integration of V2X technologies, such as RSU-assisted cooperative localization, to enhance system robustness in GNSS-denied environments. These methods offer complementary spatial cues in challenging settings like underground parking and urban canyons, potentially improving both localization and detection performance.

\bibliographystyle{IEEEtran}
\bibliography{reference}

\end{document}